\documentclass[none]{acmart}
\AtBeginDocument{%
  \providecommand\BibTeX{{%
    \normalfont B\kern-0.5em{\scshape i\kern-0.25em b}\kern-0.8em\TeX}}}

\setcopyright{none}
\copyrightyear{2023}
\acmYear{2023}
\acmDOI{XXXXXXX.XXXXXXX}

\acmJournal{TECS}
\acmVolume{1}
\acmNumber{1}
\acmArticle{1}
\acmMonth{8}

\usepackage{scalerel}
\usepackage{tikz}
\usetikzlibrary{svg.path}
\usepackage{amsmath}
\usepackage{amsfonts}
\usepackage{mathtools}
\usepackage[ruled,vlined,noend,linesnumbered]{algorithm2e}
\usepackage{caption}
\usepackage{subcaption}
\usepackage{multirow}
\usepackage{booktabs}
\usepackage{blkarray}
\usepackage[utf8]{inputenc}
\usepackage{fancyhdr}
\usepackage{enumitem}
\usepackage{xcolor}
\usepackage{textgreek}
\usepackage{hyperref}

\begin{document}

\title{BREATHE: Second-Order Gradients and Heteroscedastic Emulation based Design Space Exploration}

\author{Shikhar Tuli}
\email{stuli@princeton.edu}
\orcid{0000-0002-9230-5877}
\author{Niraj K. Jha}
\email{jha@princeton.edu}
\affiliation{%
  \institution{Princeton University}
  \streetaddress{Department of Electrical and Computer Engineering}
  \city{Princeton}
  \state{NJ}
  \country{USA}
  \postcode{08544}
}

\thanks{This work was supported by NSF Grant No. CNS-2216746.}

\renewcommand{\shortauthors}{Tuli and Jha}

\begin{abstract}
Researchers constantly strive to explore larger and more complex search spaces in various scientific studies and physical 
experiments. However, such investigations often involve sophisticated simulators or time-consuming experiments that make 
exploring and observing new design samples challenging. Previous works that target such applications are typically 
sample-inefficient and restricted to vector search spaces. To address these limitations, this work proposes a constrained 
multi-objective optimization (MOO) framework, called BREATHE, that searches not only traditional vector-based design spaces 
but also graph-based design spaces to obtain best-performing graphs. It leverages second-order gradients and actively trains 
a heteroscedastic surrogate model for sample-efficient optimization. In a single-objective vector optimization application, 
it leads to 64.1\% higher performance than the next-best baseline, random forest regression. In graph-based search, 
BREATHE outperforms the next-best baseline, i.e., a graphical version of Gaussian-process-based Bayesian optimization, with up 
to 64.9\% higher performance. In a MOO task, it achieves up to 21.9$\times$ higher hypervolume than the state-of-the-art 
method, multi-objective Bayesian optimization (MOBOpt). BREATHE also outperforms the baseline methods on most standard MOO benchmark applications.
\end{abstract}

\ccsdesc[500]{Mathematics of computing~Mathematical optimization}
\ccsdesc[500]{Computing methodologies~Machine learning}

\keywords{active learning, constrained multi-objective optimization, neural networks, surrogate modeling.}

\maketitle

\section{Introduction}

The number and complexity of applications that require search over a large design space to obtain 
well-performing solutions are increasing at a significant rate. Various scientific studies aim to search for the best 
design option in the context of diverse applications, from transistor modeling to astronomical experiments. 
For instance, modeling the effects of the geometric parameters of a nanowire field-effect transistor on hot-carrier 
injection~\cite{hci} (to study transistor reliability) requires several measurements on a diverse set of parameter 
choices~\cite{hci_modeling}. Each measurement involves the application of a linear sweep over the drain and gate 
voltages ($V_{ds} = V_{gs}$) to analyze transistor gate degradation due to \emph{hot} (energetic) carriers trapped in the 
gate dielectric. In cosmology, the likelihood of the Lyman-alpha transition~\cite{draine2010physics} provides strong 
constraints on cosmological parameters and intergalactic medium features, quantities of great interest to astrophysicists. 
Extracting the likelihood, however, requires simulations to be performed in a high-dimensional parametric 
space~\cite{lyman_alpha}. Other domains that require efficient design space exploration include neural architecture 
search (NAS)~\cite{nas_al}, cyber-physical systems~\cite{assent}, qubit design~\cite{qubit_design}, and many more. However, 
these studies either involve computationally expensive simulators or time-consuming real-world experimentation, and 
searching the entire design (or parametric) space via a brute-force approach is typically not possible. 

Several \emph{black-box} optimization methods target efficient search of a design space~\cite{bbo}. These include random 
search, regression trees, Gaussian-process-based Bayesian optimization (GP-BO), random forest regression, etc. However, 
gradient descent typically outperforms these traditional \emph{gradient-free} approaches. For instance, gradient-based 
optimization outperforms traditional methods in the domain of NAS~\cite{gradient_free_opt, flexibert, darts}. However, 
leveraging this approach requires a differentiable surrogate of the \emph{black-box} one wishes to optimize. Moreover, there 
is often a lack of knowledge and skill in machine learning (ML) among domain experts (device physicists, astronomers, etc.) 
in order to develop and optimize such surrogate models. Hence, there is a need for a plug-and-play sample-efficient 
gradient-based optimization pipeline that is applicable to diverse domains with variegated input/output constraints. 

To tackle the abovementioned challenges, we propose a novel optimization method, \underline{B}ayesian optimization using 
second-order g\underline{r}adi\underline{e}nts on an \underline{a}ctively \underline{t}rained \underline{h}eteroscedastic 
\underline{e}mulator (BREATHE). It is an easy-to-use approach for efficient search of diverse design spaces where input 
simulation, experimentation, or annotation is computationally expensive or time-consuming. BREATHE is applicable to 
both \emph{vector} and \emph{graph} optimization. We call the corresponding versions V-BREATHE and G-BREATHE, respectively. 
G-BREATHE is a novel graph-optimization approach that optimizes both the graph architecture 
and its components (node and edge weights) to maximize output performance while honoring user-defined constraints. Rigorous 
experiments demonstrate the benefits of our proposed approach over baseline methods for diverse applications.

The main contributions of the article are as follows.
\begin{itemize}
    \item We propose V-BREATHE, an efficient \emph{vector optimization} method, that is widely applicable to diverse domains. 
It leverages gradient-based optimization using backpropagation to the input (GOBI)~\cite{tuli2021cosco} implemented on a 
heteroscedastic surrogate model~\cite{npn}. It executes output optimization and supports constraints on the input or output. 
We propose the concept of \emph{legality-forcing} on gradients to support constrained optimization and leverage gradients 
in discrete search spaces. To handle output constraint violations, we use \emph{penalization} on the output. V-BREATHE requires minimal user expertise in ML.
    \item We propose G-BREATHE to apply BREATHE to graphical problems. It is a \emph{graph optimization} framework that 
searches for the best-performing graph architecture while optimizing the node and edge weights as well. It supports 
multi-dimensional node and edge weights, thus targeting a much larger set of applications than vector optimization.
    \item We further enhance V-BREATHE and G-BREATHE to support multi-objective optimization where the desired output is a 
set of \emph{non-dominated solutions} that constitute the Pareto front for a given problem. Using multiple random cold restarts, when implementing GOBI, our optimization pipeline can even tackle non-convex Pareto fronts. Our proposed 
approach achieves a considerably higher hypervolume with fewer queried samples than baseline methods.
\end{itemize}

The rest of the paper is organized as follows. Section~\ref{sec:background} presents background material on vector and graph 
optimization methods. Section~\ref{sec:methodology} illustrates the BREATHE algorithm in detail. Section~\ref{sec:exp_setup} 
describes the experimental setup and the baselines that we compare against. Section~\ref{sec:results} explains the results. Section~\ref{sec:discussion} discusses the results in more detail and points out the limitations of the proposed approach. Finally, Section~\ref{sec:conclusion} concludes the article.

\section{Background and Related Work}
\label{sec:background}

In this section, we provide background and related work on optimization using an actively-trained surrogate model.

\subsection{Vector Optimization}

We refer to the optimization of a multi-dimensional vector (say, $x \in \mathbb{R}^d$) as \emph{vector optimization}. 
Its application to a black-box function falls under the domain of \emph{black-box} optimization. Mathematically, one can 
represent this problem as follows.
\begin{equation}
\begin{aligned}
    \min \ \ &F(x) \\
    \text{s.t.} \ \ &x^L_i \le x_i \le x^U_i \ &&i = 1, \ldots, d 
\end{aligned}
\end{equation}
Here, $F$ is the black-box function we need to optimize; $x_i$ is the $i$-th variable in the design space; $x^L_i$ and 
$x^U_i$ are its lower and upper bounds. We may not have a closed-form expression for $F$ (a black-box); thus, finding a 
solution $x^*$ may not be easy.

Many works target vector optimization. Random search uniformly samples inputs within the given bounds. 
Gradient-boosted regression trees (GBRTs) model the output using a set of decision trees~\cite{lightgbm}. 
GP-BO~\cite{gp_bo} approximates performance through Gaussian process regression and optimizes an acquisition function 
through the L-BFGS method~\cite{lbfgs}. Other optimization methods leveraging a GP-based surrogate suffer from the bottlenecking $\mathbf{argmax}$ operation over the entire design space~\cite{aegis}. Random forests fit various randomized decision trees over sub-samples of the dataset. 
BOSHNAS~\cite{flexibert} searches for the best neural network (NN) architecture for an ML task. It outperforms other 
optimization techniques in the application of NAS to convolutional NNs~\cite{codebench} and transformers~\cite{flexibert}. These approaches rely on active
learning~\cite{al_survey}, in which the surrogate model, which can be a regression tree or an NN, interactively queries the 
simulator (or the experimental setup) to label new data. We use the new data to update the model at each iteration. 
This updated model forms new queries that lead to higher predicted performance. We iterate through this process until
it meets a convergence criterion. Finally, this yields the input with the best output performance.

The abovementioned approaches do not consider optimization under constraints. Optimizing the objective function while adhering 
to user-defined constraints falls under the domain of constrained optimization. Mathematically,
\begin{equation}
\begin{aligned}
    \min \ \  &F(x) \\
    \text{s.t.} \ \ &x^L_i \le x_i \le x^U_i \ &&i = 1, \ldots, d \\
                  &C^I_j(x) \le 0 \ &&j = 1, \ldots, J' \\
                  &C^E_k(x) = 0 \ &&k = 1, \ldots, K'
\end{aligned}
\end{equation}
for $J'$ inequality constraints and $K'$ equality constraints. One can convert each equality constraint to two inequality 
constraints. Thus, we can simplify the above problem as follows:
\begin{equation}
\begin{aligned}
\label{eq:vector_constrained_soo}
    \min \ \  &F(x) \\
    \text{s.t.} \ \ &x^L_i \le x_i \le x^U_i \ &&i = 1, \ldots, d \\
                  &C_j(x) \le 0 \ &&j = 1, \ldots, J
\end{aligned}
\end{equation}
where $J = J' + 2 K'$.

This problem belongs to the class of \emph{constrained} single-objective optimization (SOO) problems. One could also search 
the input space to optimize multiple objectives simultaneously while honoring the input constraints. We refer to this
class of problems as \emph{constrained} multi-objective optimization (MOO) problems~\cite{deb2011multi}. Mathematically,
\begin{equation} 
\begin{aligned}
\label{eq:vector_constrained_moo}
    \min \ \  &F_m(x) \ &&m = 1, \ldots, M \\
    \text{s.t.} \ \ &x^L_i \le x_i \le x^U_i \ &&i = 1, \ldots, d \\
                  &C_j(x) \le 0 \ &&j = 1, \ldots, J
\end{aligned}
\end{equation}
for $M$ objective functions.

Previous works propose various methods to solve MOO problems.  
Non-dominated sorting genetic algorithm-2 (NSGA-2)~\cite{nsga2} is a seminal evolutionary algorithm (EA)-based optimization method that evolves a set of candidates 
across generations into better-performing solutions. Multi-objective evolutionary algorithm based on decomposition 
(MOEA/D)~\cite{moead} is another popular EA-based method that decomposes 
a MOO problem into multiple SOO problems and optimizes them simultaneously. \textcolor{black}{Many state-of-the-art search techniques are based on evolutionary methods~\cite{chugh2019survey, rahi_steady_state}. Since the proposed approach is a surrogate-based method, we only compare it against the \emph{representative} EA-based methods 
mentioned above.} Multi-objective regionalized Bayesian optimization (MORBO)~\cite{morbo} is a MOO framework based on Bayesian 
optimization (BO). It performs BO in multiple local regions of the design space to identify the global optimum. MOBOpt~\cite{mobopt} is yet another surrogate-based Bayesian optimization framework for MOO problems. The proposed V-BREATHE algorithm solves both SOO and MOO problems.

\subsection{Graph Optimization}

In the above scenario, input $x \in \mathbb{R}^d$ is a vector. However, in many applications, $x$ may be a graph, i.e., 
$x \in \mathcal{G}$, where $\mathcal{G}$ is the space of \emph{legal} graphs (see Section~\ref{sec:methodology_const}). In 
this scenario, \emph{graph optimization} refers to searching for an input graph that optimizes (single or) multiple output 
objectives under given constraints. Mathematically, 
\begin{equation} 
\begin{aligned}
\label{eq:graph_constrained_moo}
    \min \ \  &F_m(x) \ &&m = 1, \ldots, M \\
    \text{s.t.} \ \ &x \in \mathcal{G} \\
                  &C_j(x) \le 0 \ &&j = 1, \ldots, J
\end{aligned}
\end{equation}
where we define $\mathcal{G}$ based on a set of legal node connections (edges) along with node and edge weight bounds.

Traditional works on \emph{graph optimization} target specific problems, such as max-flow/min-cut~\cite{max_flow_min_cut}, 
graph partitioning~\cite{graph_partitioning}, graph coloring~\cite{graph_coloring}, routing~\cite{tsp}, etc. However, these 
optimization problems aim to either find a subset of a given graph or annotate a given graph. In this work, we target an 
orthogonal problem: find the best-performing graph for the given objective function(s). This involves searching for the set 
of nodes (or vertices) and edges along with their weights. Existing works solve this problem with limited scope, i.e., they may not consider all graph constraints~\cite{assent} (when converting the problem into vector optimization) or are only applicable to NN models~\cite{nas_al}. Moreover, graph optimization involves searching for not only the node/edge values (which could be represented as multi-dimensional vectors) but also the connections. Directly \emph{flattening} a graph and implementing vector optimization methods does not perform well, as we show in this work, as such methods would not be able to look for new connections that modify the graph architecture. This calls for novel search techniques that directly implement optimization on graphical input. The proposed G-BREATHE algorithm solves both SOO and MOO 
graph problems.

\section{Methodology}
\label{sec:methodology}

\begin{figure}
    \centering
    \includegraphics[width=0.95\linewidth]{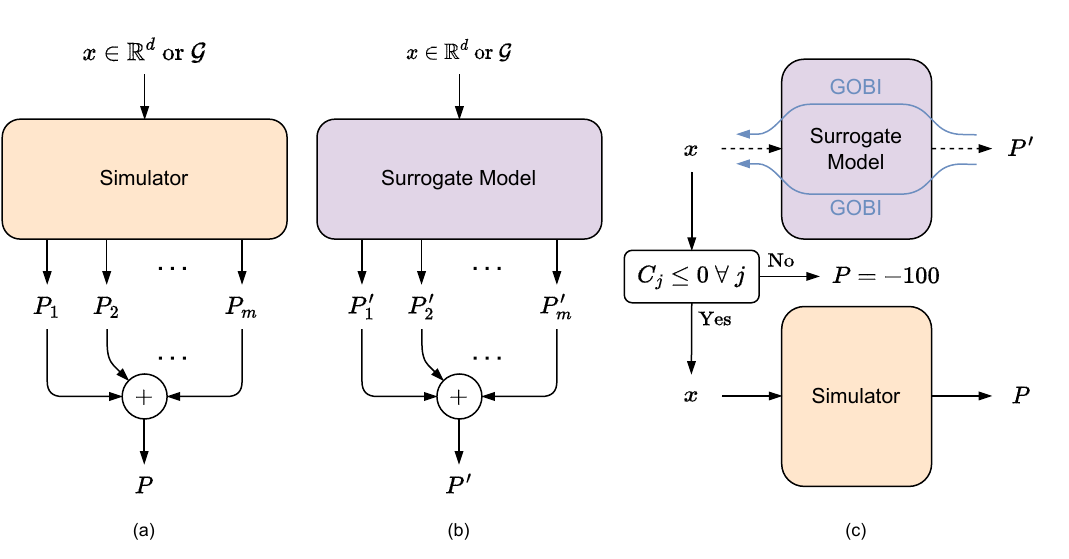}
    \caption{Overview of the BREATHE framework. (a) The simulator takes an input $x$, which can be a vector (in $\mathbb{R}^d$) 
or a graph (in $\mathcal{G}$). It outputs the performance value corresponding to each objective ($P_i$). $P$ is a convex 
combination of individual performance measures for SOO. (b) The surrogate model is a \emph{lightweight} (i.e., computationally inexpensive relative to the simulator) and 
\emph{differentiable} emulator that mimics the simulator; its predicted performance is $P'$. (c) GOBI is applied to the 
surrogate model to obtain $x$ with a higher predicted performance $P'$. The dataset is updated with the simulated value 
$P$ to enable retraining of the surrogate model.}
    \label{fig:breathe_framework}
\end{figure}

In this section, we discuss the BREATHE framework that leverages GOBI on a heteroscedastic surrogate model. 
Fig.~\ref{fig:breathe_framework} summarizes the BREATHE framework.

\subsection{V-BREATHE}
\label{sec:vbreathe}

V-BREATHE is a widely-applicable vector-input-based optimizer that runs second-order gradients on a \emph{lightweight} 
NN-based surrogate model to predict not only the output objective value but also its epistemic and aleatoric uncertainties. 
It leverages an active-learning framework to optimize the upper confidence bound (UCB) estimate of the output objective. 
In this context, we freeze the model weights and backpropagate the gradients to update the \emph{input} (not the model weights) to optimize the output objective. 
We then query the simulator to obtain the output of the new queried sample and retrain the surrogate model. This iterative 
search process continues until convergence. We describe the V-BREATHE optimizer in detail next.

\subsubsection{Uncertainty Types}

Prediction uncertainty may arise from not only the approximations made in the surrogate modeling process or limited data and 
knowledge of the design space but also the natural stochasticity in observations. The former is termed \emph{epistemic} 
uncertainty and the latter \emph{aleatoric} uncertainty. The epistemic uncertainty, also called reducible uncertainty, arises from a lack of knowledge or information, and the aleatoric uncertainty, also called irreducible uncertainty, refers to the inherent variation in the 
system to be modeled.

In addition, uncertainty in output observations may also be 
data-dependent (known as \emph{heteroscedastic} uncertainty). Accounting for such uncertainties in the optimization objective 
requires a surrogate that also models them.

\subsubsection{Surrogate Model}

Following the surrogate modeling approach used in BOSHNAS~\cite{flexibert}, a state-of-the-art NAS framework, we model the 
output objective and aleatoric uncertainty using a natural parameter network (NPN)~\cite{npn} $f(x; \theta)$. We model the 
epistemic uncertainty using a teacher network $g(x; \theta')$ and its student network $h(x; \theta'')$. Here, $\theta$, 
$\theta'$, and $\theta''$ refer to the trainable parameters of the respective models. We leverage GOBI on $h$ to avoid 
numerical gradients due to their poor performance~\cite{flexibert}. We have $(\mu, \sigma) \leftarrow f(x; \theta)$, where 
$\mu$ is the predicted output objective (i.e., a surrogate of $F$) and $\sigma$ is the aleatoric uncertainty. Moreover, $h$ 
predicts a surrogate ($\hat{\xi}$) of the epistemic uncertainty ($\xi$). The teacher network $g$ models the epistemic 
uncertainty via Monte Carlo dropout~\cite{flexibert}.

We model the output objective in the $[0, 1]$ interval for easier convergence. We implement this in the surrogate model by 
feeding the output to a sigmoid activation. To implement this, we normalize the output objective $F$ with respect to its 
maximum permissible value and maximize the performance measure:
\begin{equation} 
\label{eq:perf}
    P = 1 - \frac{F}{F^{MAX}}, \quad \ F^{MAX} = M_O \max_{x, \forall (x, P) \in \Delta} F
\end{equation}
where $\Delta$ is the set of currently observed samples in the design space and $M_O \ge 1$ is a multiplicative overhead 
factor. If we observe a larger value during the search process, we re-annotate the observed data with the updated value 
of $F^{MAX}$ and retrain the \emph{lightweight} surrogate model.

\subsubsection{Active Learning and Optimization}

To use GOBI and obtain queries that perform well, we initialize the surrogate model by training it on a 
randomly sampled set of points in the design space. We call this set the seed dataset.
To effectively explore globally optimal design points, the seed dataset should be as representative of the 
design space as possible. For this, we use low-discrepancy sequence sampling 
strategies~\cite{random_number_generation}. Specifically, V-BREATHE uses Latin hypercube sampling to 
obtain divergent points in its sampled set \textcolor{black}{(parallel works show that this indeed performs better 
than other low-discrepancy sampling methods in maximizing the diversity of the sampled points~\cite{edgetran})}. We evaluate these $N_{\Delta_0}$ initial samples using the 
(albeit expensive) simulator and train the surrogate model on this seed dataset $\Delta_0$. 
Then, we run second-order optimization on 
\begin{equation} \label{eq:ucb}
    \text{UCB} = \mu + k_1 \cdot \sigma + k_2 \cdot \hat{\xi}    
\end{equation}
where $k_1$ and $k_2$ are hyperparameters. We employ the UCB estimate instead of other acquisition functions as it results in the fastest convergence as per previous works that leverage gradient-based optimization~\cite{flexibert, codebench}. Nevertheless, we leave the application of other acquisition functions to future work.

\subsubsection{Incorporating Constraints}
\label{sec:methodology_const}

Since one cannot directly add symbolic constraints to an NN, we train the surrogate model with a 
sample with very low performance value that does not satisfy the output constraints (also called \emph{penalization}). For instance, if an output $F_m$ does not meet a constraint (say, $C_{j'} > 0$), we set the corresponding 
performance to $P = -100$, which should otherwise be in the $[0, 1]$ interval. This forces the surrogate model to learn 
the distribution of input samples that do not satisfy the desired output constraints. 

\begin{figure}
    \centering
    \includegraphics[width=0.5\linewidth]{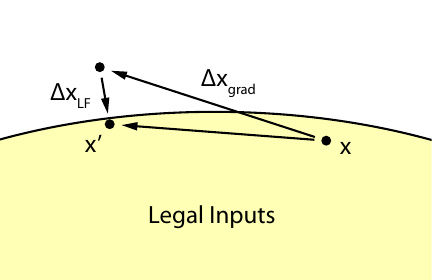}
    \caption{Schematic illustration of \emph{legality-forcing}. GOBI results in the gradient step $\Delta x_{\text{grad}}$ that is outside the set of \emph{legal} inputs (which may be a subset of the design space). $\Delta x_{\text{LF}}$ implements a step forcing the resultant input to be \emph{legal}.}
    \label{fig:legality_forcing}
\end{figure}

While running GOBI, if the updated input at an epoch does not satisfy the constraints (i.e., is 
an \emph{illegal} input), we set it to the nearest \emph{legal} input \textcolor{black}{(based on Euclidean distance)} that satisfies the constraints. 
One can consider this as adding an additional \emph{force} while running gradient descent in the input 
space that iteratively makes the updated input \emph{legal}. We call this approach \emph{legality-forcing}. Fig.~\ref{fig:legality_forcing} explains this using a working schematic.

\subsubsection{Simultaneous Optimization of Multiple Objectives}
\label{sec:methodology_moo}

To support MOO with V-BREATHE, we need to run GOBI to obtain new input queries that optimize all $F_m$ 
in Eq.~(\ref{eq:vector_constrained_moo}). To tackle this problem, one could try to optimize a sum of all 
objectives. However, each point on the Pareto front weights each objective differently. Hence, we optimize 
multiple convex combinations of $F_m$'s. More concretely,
\begin{equation} \label{eq:perf_moo}
\begin{aligned}
    P = \sum_m \alpha_m P_m, \quad m = 1, \ldots, M 
\end{aligned}
\end{equation}
where $P_m$ is a function of $F_m$ as per Eq.~(\ref{eq:perf}) and $\alpha_m$'s, where 
$\sum_m \alpha_m = 1$, are hyperparameters that determine the weight assigned to each objective. Thus, 
different samples of these weights would result in different \emph{non-dominated solutions}, i.e., points 
on the Pareto front. In this context, V-BREATHE maximizes the performance measure $P$ for every set of 
weight samples using GOBI.

\subsection{G-BREATHE}
\label{sec:gbreathe}

G-BREATHE implements the V-BREATHE algorithm on graphical input. Instead of using a fully-connected NN, 
which assumes a vector input, we use a graph transformer~\cite{graph_txf} network as our surrogate model. We use a graph transformer as it is a state-of-the-art model for graphical input.
We characterize the input graph by its nodes, edges, and multi-dimensional weight values. 

While running GOBI, we backpropagate the gradients to the node and edge weights. If all the edge weights 
fall below a threshold ($\epsilon_E = 10^{-5}$ for non-binary edge weights and $\epsilon_E = 0.5$ for binary edge weights), we remove that edge. To explore diverse graphs with varied node/edge combinations 
and weights, we sample randomly generated graphs at each iteration of the search process and run GOBI on 
the trained surrogate to look for better versions of those graphs (i.e., with the same connections but 
different node/edge weights that maximize performance). The rest of the setup is identical to that of 
V-BREATHE. From now on, we will use the term BREATHE to refer to either V-BREATHE or G-BREATHE based on 
the input type unless otherwise specified.

\SetKwComment{Comment}{/* }{ */}
\SetKw{Kwis}{is}
\SetKwFunction{KwGOBI}{\textbf{GOBI}}
\SetKwFunction{KwFit}{\textbf{fit}}
\SetKwFunction{KwSimulate}{\textbf{simulate}}
\SetKwFunction{KwArgmax}{\textbf{argmax}}
\SetKwFunction{KwRandom}{\textbf{random}}
\begin{algorithm}[t]
\DontPrintSemicolon
\SetAlgoLined
\KwResult{\textbf{best} $x$}
\textbf{Initialize:} convergence criterion, uncertainty sampling 
prob. ($\alpha$), diversity sampling prob. ($\beta$), surrogate model ($f$, $g$, and 
$h$), initial dataset $\Delta_0$, design space $\mathcal{D}$, $i = 0$; \par
\While{convergence criterion not met}{
  \uIf{prob. $\sim U(0,1) < 1 - \alpha - \beta$}{
   $\KwFit$($f$, $g$, $h$; $\Delta_i$); \par \label{line:fit}
   $x$ $\gets$ $\KwGOBI$($f$, $h$); \Comment*{Optimization step} \label{line:opt}
   $P \gets \KwSimulate$($x$); \label{line:simulate}
  }
  \uElse{
   \uIf{$1 - \alpha - \beta \le$ prob. $< 1 - \beta$}{
    $x$ $\gets$ $\KwArgmax$($k_1 \cdot \sigma + k_2 \cdot \hat{\xi}$); \label{line:uncertainty} \Comment*{Uncertainty sampling} 
   }
   \uElse{
    $x \gets \KwRandom$($\mathcal{D}$); \label{line:diversity} \par
    \Comment*{Diversity sampling}
   }
   $C \gets$ True; \par \label{}
   \For{$j = 1$ \KwTo $J$}{
    \If{$C_j(x) > 0$}{
     $C \gets$ False; \Comment*{Output constraint not satisfied}
     $\textbf{break}$;
    }
   }
   \uIf{$C$ \Kwis \textup{True}}{
    $P \gets \KwSimulate$($x$);  
   }
   \uElse{
    $P \gets -100$; \label{line:const_not_satisfied}
   }
  }
  $i \gets i + 1$; \par
  $\Delta_i \gets (x, P) \cup \Delta_{i-1}$; \par
 }
 \caption{BREATHE} 
 \label{alg:breathe}
\end{algorithm}

Algorithm~\ref{alg:breathe} summarizes the BREATHE algorithm \textcolor{black}{(for both SOO and MOO settings)}. Starting from an initial seed dataset 
$\Delta_0$, we run the following steps until convergence. To trade off between exploration and 
exploitation, we consider two probabilities: uncertainty-based exploration ($\alpha$) and diversity-based 
exploration ($\beta$). With probability $1 - \alpha - \beta$, we run second-order GOBI using the surrogate 
model to maximize the UCB in Eq.~(\ref{eq:ucb}). To achieve this, we first train the surrogate model 
(a combination of $f$, $g$, and $h$) on the current dataset (line~\ref{line:fit}). \textcolor{black}{For SOO,
we initialize only one surrogate model, however, for MOO, we initialize separate surrogate models for each
set of randomly initialized $\alpha_m$'s.} We then generate a new 
query point by freezing the weights of $f$ and $h$ and run GOBI along with \emph{legality-forcing} to obtain 
an $x$ (adhering to constraints) with better-predicted performance (line~\ref{line:opt}). We then simulate 
the obtained query to obtain the performance measure $P$ (line~\ref{line:simulate}). For SOO, this 
corresponds to a single $\alpha_1 = 1$, but for MOO, this corresponds to a random selection of $\alpha_m$'s, 
and we obtain $P$ as per Eq.~(\ref{eq:perf_moo}). With $\alpha$ probability, we sample the search space 
using the combination of aleatoric and epistemic uncertainties, $k_1 \cdot \sigma + k_2 \cdot \hat{\xi}$, 
to find a point where the performance estimate is the most uncertain (line~\ref{line:uncertainty}). We also 
choose a random point with probability $\beta$ (line~\ref{line:diversity}) to avoid getting stuck in a 
localized search subset. The exploration steps also aid in reducing the inaccuracies in surrogate modeling for unexplored regions of the design space. Finally, we run the simulation only if the obtained input adheres to user-defined 
constraints ($C_j$'s in Eqs.~(\ref{eq:vector_constrained_soo}), (\ref{eq:vector_constrained_moo}), and 
(\ref{eq:graph_constrained_moo})), and penalize the output performance by setting $P = -100$ otherwise (line~\ref{line:const_not_satisfied}).

\section{Experimental Setup}
\label{sec:exp_setup}

In this section, we present the setup behind various experiments, including the applications for vector and 
graph optimization, baselines for comparison, and details of the surrogate models.

\subsection{Evaluation Applications}

The applications with which we test our optimizers include two in the domain of vector optimization: operational amplifier (op-amp) 
and waste-water treatment plant (WWTP), and two in the domain of graph optimization: smart home and network. We also test the V-BREATHE algorithm against previously-proposed MOO baselines on standard benchmarks.
Finally, we use the problem of synchronous optimal pulse-width modulation (SO-PWM) of three-level inverters 
for scalability analysis. We describe each application in detail next.

\subsubsection{Op-amp} 

\begin{figure}[t]
    \centering
    \includegraphics[width=0.9\linewidth]{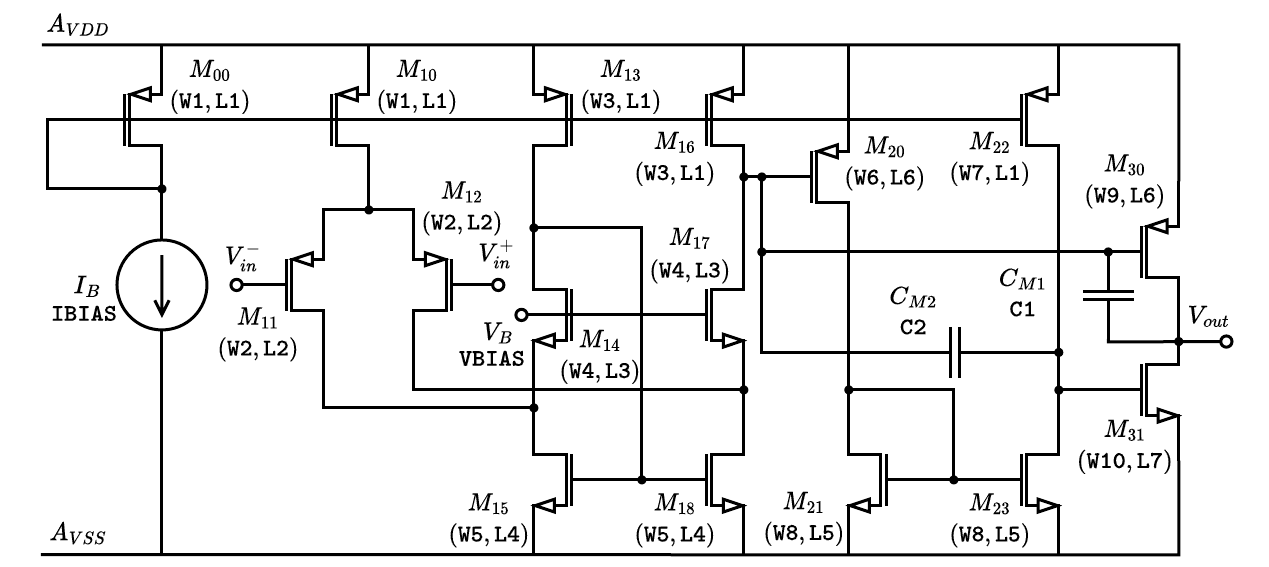}
    \caption{Circuit diagram of the three-stage op-amp. Variables that constitute the input design space are presented in typewriter font.}
    \label{fig:circuit_diagram}
\end{figure}

This application involves the optimization of a three-stage op-amp featuring a 
positive-feedback frequency compensation technique~\cite{opamp_freq}. It has 24 input variables 
(supply/bias voltages and currents, load resistance, load capacitance, transistor widths and lengths, 
etc.), three constraints (lower/upper limits on DC gain, phase margin, and unity gain frequency), and one 
optimization objective (minimization of power consumption). We implement the simulations using the Cadence 
Spectre circuit simulator~\cite{spectre}.

Fig.~\ref{fig:circuit_diagram} shows a circuit diagram of the op-amp along with the variables involved in the optimization process. $M_{00}$ is the bias metal-oxide semiconductor field-effect transistor (MOSFET). $M_{10}$, $M_{11}$, and $M_{12}$ are part of the differential pair. $M_{13}$-$M_{18}$ are part of the folded cascode. MOSFETs $M_{20}$-$M_{23}$ constitute the second stage, while $M_{30}$ and $M_{31}$ constitute the third stage of the op-amp. We use two compensation capacitors: $C_{M1}$ and $C_{M2}$. The input optimization space comprises 24 variables. These include 10 MOSFET widths ($\texttt{W1}$-$\texttt{W10}$), seven MOSFET lengths ($\texttt{L1}$-$\texttt{L7}$), capacitor values ($\texttt{C1}$ and $\texttt{C2}$), bias voltage ($\texttt{VBIAS}$), bias current ($\texttt{IBIAS}$), load resistance ($\texttt{RLOAD}$), load capacitance ($\texttt{CLOAD}$), and supply voltage ($\texttt{VSUPPLY}$).

We constrain the DC gain to be greater than 68, the phase margin to be between 31.8\textdegree and 130.0\textdegree, and the 
unity gain frequency to be greater than 1.15 MHz~\cite{asco}.

\subsubsection{WWTP} 

Waste-water treatment removes contaminants and converts waste water into an effluent that one can 
return to the water cycle. Specifically, we consider the four-stage Bardenpho process~\cite{bardenpho}. 
Optimizing the WWTP involves 14 input variables (flow split percentages along with reactor volumes, 
temperatures, and dissolved oxygen levels) and four output objectives [fractions of chemical oxygen 
demand (COD), namely $S_I$: inert soluble, $S_S$: readily biodegradable, $X_I$: inert particulate, and 
$X_S$: slowly biodegradable compounds]. For SOO, we optimize the net COD as a sum, i.e., 
$S_I + S_S + X_I + X_S$. However, for MOO, we optimize all objectives simultaneously to obtain 
commonly-studied Pareto fronts: $S_I$ vs. $S_S$ and $X_I$ vs. $X_S$. We use a publicly available 
simulation software\footnote{The WWTP simulator is available at 
\url{https://github.com/toogad/PooPyLab_Project}} for evaluating different WWTPs.

\begin{figure*}
    \centering
    \includegraphics[width=\linewidth]{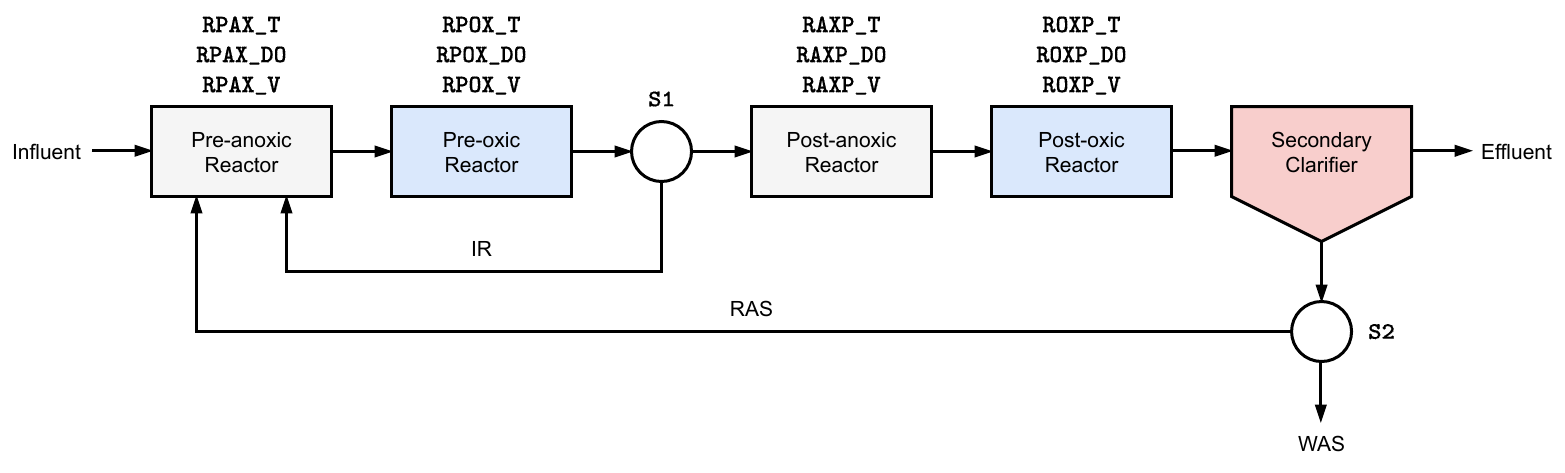}
    \caption{Simplified schematic of the four-stage Bardenpho process. Variables that constitute the input design space are presented in typewriter font.}
    \label{fig:bardenpho}
\end{figure*}

Fig.~\ref{fig:bardenpho} shows a simplified schematic of the four-stage Bardenpho process. It consists of four reactors
and a secondary clarifier. There are three design parameters for each reactor, namely, the maintained temperature (in
\textdegree C), the dissolved oxygen (DO, in mg/L), and the reactor volume (in m$^3$). The first split ($\texttt{S1} \in
[1, 8]$) determines the amount of sludge for internal re-circulation (IR). The second split ($\texttt{S2} \in [0, 1]$)
determines the ratio of return-activated sludge (RAS) and waste-activated sludge (WAS). This results in 14 input
variables for optimization. We limit the temperatures in the range 5-15 \textdegree C, volumes in 100-15000 m$^3$, and
DO in 0-5 mg/L. These ranges are typically used in many plants.

\subsubsection{Smart Home}

A smart home consists of multiple Internet-of-Things (IoT) devices. These include smart doorbells, 
smart locks, entry/exit sensors, smart door cameras, home assistants, smart thermostats, etc. We support 
different network types, namely WiFi, Zigbee~\cite{zigbee}, and Z-Wave~\cite{zwave}. Further, we need to 
connect different room types (hall, master bedroom, bedrooms, entry hall, living room, kitchen/dining area, 
and outside porch). Each room can also have multiple IoT devices. The task is to design a smart home, with 
constraints on the total number of windows and the total area, that minimizes the number of cyber-physical 
attacks (one optimization objective). For instance, an attacker could use a light command to maliciously 
manipulate devices in rooms with windows (given that the devices use Zigbee or Z-Wave and have a clear 
line of sight through the windows in the room). We used a Python-based simulator\footnote{The smart home 
graphical simulator is available at \url{https://github.com/rahulaVT/SIM_app}} for testing our proposed 
optimizer.

We now describe the graph formulation of a smart home in G-BREATHE. There can be a total of 9 nodes, one for each room type, except bedrooms, which can be three in number. We represent each node (or room) by a 12-dimensional weight vector consisting of categorical and continuous features. The weight vector includes the room type, room area, number of windows, number of each IoT device (we support a total of eight devices), and the type of network used. We restrict all the devices in a smart home to only one network type (although the WiFi router always connects to the gateway via a WiFi connection). An edge weight is a Boolean value, i.e., whether the two nodes are connected or not. There are three constraints: the total number of windows should be more than or equal to three, the total area of the smart home should be greater than 300 m$^2$ and less than 600 m$^2$, and the network connection for all IoT devices should be the same. Table~\ref{tbl:smart_home_design_space} summarizes the design parameters for the smart home application.

\begin{table*}[]
\caption{Design parameters associated with the smart home application. The table shows the number of rooms allowed in the graph along with their area limits. It also shows the total number of IoT devices allowed in a smart home for each device type 
(S.: Smart).}
\centering
\resizebox{\linewidth}{!}{
\begin{tabular}{@{}lcccccccc@{}}
\toprule
\multicolumn{9}{c}{\textbf{Room Types}} \\ \midrule
\multicolumn{1}{l|}{\textbf{Design Space Parameter}} & \textbf{Hall} & \textbf{Master Bedroom} & \multicolumn{2}{c}{\textbf{Bedroom}} & \textbf{Entry Hall} & \textbf{Living Room} & \textbf{Kitchen/Dining} & \textbf{Porch} \\ \midrule
\multicolumn{1}{l|}{Number} & 1 & 1 & \multicolumn{2}{c}{1-3} & 1 & 1 & 1 & 1 \\ [1mm]
\multicolumn{1}{l|}{Area (m$^2$)} & 10-50 & 24-100 & \multicolumn{2}{c}{10-50} & 10-40 & 50-120 & 50-100 & 80-120 \\ \midrule
\multicolumn{9}{c}{\textbf{IoT Devices}} \\ \midrule
\multicolumn{1}{l|}{\textbf{Design Space Parameter}} & \textbf{WiFi Modem} & \textbf{S. Doorbell} & \textbf{Gateway} & \textbf{S. Lock} & \textbf{Entry Sensor} & \textbf{S. Door Camera} & \textbf{Home Assistant} & \textbf{S. Thermostat} \\ \midrule
\multicolumn{1}{l|}{Number} & 1 & 1 & 1 & 1 & 1-3 & 1 & 1-8 & 1-3 \\ \bottomrule
\end{tabular}}
\label{tbl:smart_home_design_space}
\end{table*}

\subsubsection{Network}

Higher bandwidth and lower latency connections available in modern networks~\cite{gill2019transformative} 
have enabled the network edge to execute substantially more computations. However, the simulation of urban 
mobility (SUMO) domain~\cite{sumo} demands computationally-expensive simulations that must run frequently 
to ensure stable connections. In this application, we optimize the number of switches and their connections 
with data sources and mobile/edge data sinks (thus, forming a network graph) to maximize bandwidth and 
minimize network operation costs.

Each graph may have up to 25 nodes (five data sources, five data sinks, and up to 15 switches). The node weight represents the type of node: data source/sink or switch. Edges between nodes represent the bandwidth of the corresponding network connection (restricted from 128 MB/s to 1024 MB/s). As explained above, the two optimization objectives are network bandwidth and operation cost. 

We set the cost of a data source (typically a cloud server) to \$5000 while that of sink to \$1000 (an edge device). The
cost of a switch is \$1000. Adding a connection to the switch incurs an additional base cost of \$200, which increases
with bandwidth [at the rate of 0.05 \$/(MB/s)]. These costs are typical of common devices employed in networking applications.

\subsubsection{SO-PWM of Three-level Inverters}

Multi-level inverters reduce the total harmonic distortion (THD) in their alternating current (AC) output. 
SO-PWM control permits setting the maximum switching frequency to a low value without compromising on the 
THD~\cite{sopwm} of the AC output. This application has 25 inputs, 24 constraints, and two optimization 
objectives. We perform scalability tests in Section~\ref{sec:scalability} where we study the effect on 
best-achieved performance and the number of evaluation queries as we increase the number of tunable inputs 
or constraints. We implement an adapted version of the MATLAB-based simulator, available from a benchmarking 
suite~\cite{benchmark_suite}, for scalability analysis.

We represent the SO-PWM problem~\cite{sopwm} in the form of Eq.~(\ref{eq:vector_constrained_moo}) as follows:
\begin{equation} 
\begin{aligned}
\label{eq:sopwm_moo}
    \min \ \  &F_1(x) = \frac{\sqrt{\sum_k{k^{-4} \sum_{i=1}^N S(i) \cos^2{(k x_i)}}}}{\sqrt{\sum_k{k^{-4}}}} \\
    \min \ \  &F_2(x) = \left(0.32 - \sum_{i=1}^N{S(i) \cos{(x_i)}} \right)^2 \\
    \text{s.t.} \ \ &0 < x_i < \frac{\pi}{2}, \qquad\qquad\quad \ \ i = 1, \ldots, N \  \\
                    &x_{j+1} - x_j - 10^{-5} > 0,  \quad \, j = 1, \ldots, N -1
\end{aligned}
\end{equation}
where, $k = 5, 7, 11, 13, \ldots, 97$, $N = 25$, and $S(i) = (-1)^{i+1}$.

\subsubsection{Benchmark Applications}

We compare V-BREATHE against baseline methods on standard benchmarks. These include the ZDT problem suite~\cite{zdt}, the Binh and Korn (BNH) benchmark~\cite{bnh}, the Osyczka and Kundu (OSY) benchmark~\cite{osy}, and the Tanaka (TNK) benchmark~\cite{tnk}. Although these benchmarks are implemented with mathematical formulas that are easy to evaluate, to test the efficacy of various methods in low-data regimes (in the context of a computationally expensive simulator), we start with a low number of randomly sampled points (i.e., 64 in our experiments) in the seed dataset $\Delta_0$.

Table~\ref{tbl:eval_apps} summarizes the dimensions involved in each optimization 
application.

\begin{table}[]
\caption{Dimensionality of inputs, constraints, and outputs for targeted applications.}
\centering
\resizebox{0.6\columnwidth}{!}{
\begin{tabular}{@{}lccccc@{}}
\toprule
\multicolumn{6}{c}{\textbf{Vector Optimization}}                                                                                               \\ \midrule
\multicolumn{1}{l|}{\textbf{Application}} & \multicolumn{3}{c}{\textbf{Inputs}}                      & \textbf{Constraints} & \textbf{Outputs} \\ \midrule
\multicolumn{1}{l|}{Op-amp}               & \multicolumn{3}{c}{24}                                   & 3                    & 1                \\ [1mm]
\multicolumn{1}{l|}{WWTP}                 & \multicolumn{3}{c}{14}                                   & 0                    & 4                \\ [1mm]
\multicolumn{1}{l|}{SO-PWM}               & \multicolumn{3}{c}{25}                                   & 24                   & 2                \\ \midrule
\multicolumn{1}{l|}{ZDT1}                 & \multicolumn{3}{c}{30}                                   & 0                    & 2                \\ 
\multicolumn{1}{l|}{ZDT2}                 & \multicolumn{3}{c}{30}                                   & 0                    & 2                \\ 
\multicolumn{1}{l|}{ZDT3}                 & \multicolumn{3}{c}{30}                                   & 0                    & 2                \\ 
\multicolumn{1}{l|}{ZDT4}                 & \multicolumn{3}{c}{10}                                   & 0                    & 2                \\ 
\multicolumn{1}{l|}{ZDT5}                 & \multicolumn{3}{c}{80}                                   & 0                    & 2                \\ 
\multicolumn{1}{l|}{ZDT6}                 & \multicolumn{3}{c}{10}                                   & 0                    & 2                \\
\multicolumn{1}{l|}{BNH}                 & \multicolumn{3}{c}{2}                                   & 2                    & 2                \\ 
\multicolumn{1}{l|}{OSY}                 & \multicolumn{3}{c}{6}                                   & 6                    & 2                \\ 
\multicolumn{1}{l|}{TNK}                 & \multicolumn{3}{c}{2}                                   & 2                    & 2                \\ \midrule
\multicolumn{6}{c}{\textbf{Graph Optimization}}                                                                                                \\ \midrule
\multicolumn{1}{l|}{\textbf{Application}} & \textbf{Nodes} & \textbf{Node dim.} & \textbf{Edge dim.} & \textbf{Constraints} & \textbf{Outputs} \\ \midrule
\multicolumn{1}{l|}{Smart Home}           & 9              & 12                 & 1                  & 3                    & 1                \\ [1mm]
\multicolumn{1}{l|}{Network}              & 25             & 1                  & 1                  & 0                    & 2                \\ \bottomrule
\end{tabular}}
\label{tbl:eval_apps}
\end{table}

\subsection{Surrogate Models}

We now present details of the architectural decisions for the surrogate models along with the 
hyperparameters used in the BREATHE algorithm. For vector optimization, in all three surrogate models 
$f$, $g$, and $h$, we pass the input through two fully-connected hidden layers with 64 and 32 
neurons, respectively. For graph optimization, we pass the input through a graph transformer 
layer~\cite{graph_txf} with four attention heads, each with a hidden dimension of 16. We then pass the 
output of the transformer layer to a set of fully-connected layers as above. We show other hyperparameter 
choices for Algorithm~\ref{alg:breathe}, which we obtained through grid search, in Table~\ref{tbl:hyperparam}.

\textcolor{black}{Training the surrogate model on the initial dataset, $\Delta_0$, for five epochs takes about 300-400 ms on an NVIDIA A100 GPU with a batch size of 64.  This is negligible compared to the time taken by the 
simulator on a single query, e.g., hundreds of seconds (or more) for some applications.} 

\textcolor{black}{The execution time of the proposed algorithm does not increase with the number of inputs. As the number of inputs increases, the input neurons in the surrogate model increase. However, since the neural network computation is performed in parallel, the execution time remains constant. }

\begin{table}[]
\caption{Selected hyperparameters for the BREATHE algorithm.}
\resizebox{0.6\columnwidth}{!}{
\begin{tabular}{@{\hskip 0.4in}l@{\hskip 0.8in}|@{\hskip 0.8in}l@{\hskip 0.4in}}
\toprule
\textbf{Hyperparmeters}    & \textbf{Value}    \\ \midrule
$M_O$             & 1.2      \\ [1mm]
$N_{\Delta_0}$    & 64       \\ [1mm]
$k_1$, $k_2$      & 0.5, 0.5 \\ [1mm]
$\alpha$, $\beta$ & 0.1, 0.1 \\ \bottomrule
\end{tabular}}
\label{tbl:hyperparam}
\end{table}

\subsection{Baselines}
\label{sec:baselines}

For SOO with vector input, we compare BREATHE against random sampling (Random), random forest regression 
(Forest), GBRT, and GP-BO. We implement these baselines using the scikit-learn library~\cite{scikit-learn} 
with default parameters. This implies that for random forest regression, we use 100 trees, the Gini index to measure the quality of splits, and the minimum number of samples per split set to two. The GBRT optimizer uses the squared-error loss with a learning rate of 0.1, 100 boosting stages, and the minimum number of samples per split set to two. For MOO, we compare the MOO version of the proposed algorithm with 
state-of-the-art and popular baselines: NSGA-2, MOEA/D, and MOBOpt. We use the implementation from 
PyMOO~\cite{pymoo} to run NSGA-2 and MOEA/D in Python. \textcolor{black}{We use the default hyperparameters (in the PyMOO library) for these methods.} For MOBOpt, we use the source code\footnote{Source code for MOBOpt is available at: https://github.com/ppgaluzio/MOBOpt.} with default parameters.

For graph optimization, we compare G-BREATHE against graphical adaptations of the above baselines. To implement this, we
randomly generate graphs and only optimize the node/edge weights of the graph by flattening it into a vector and feeding
the vector into the vector optimization baseline. We let the baseline search the node/edge weight space for 32 iterations 
before generating new graphs. This enables these baselines to search in both spaces: graph architecture and node/edge weight 
values.

\section{Results}
\label{sec:results}

In this section, we present experimental results and comparisons of the proposed BREATHE optimizer 
with relevant baselines.

\subsection{Single-objective Vector Optimization using BREATHE}

For SOO with BREATHE and baselines approaches, we maximize the performance measure $P$ defined in 
Eq.~(\ref{eq:perf_moo}) even if the application has multiple objectives. Unlike MOO, here we choose a 
fixed combination of $\alpha_m$. Only one output objective is associated with op-amp optimization, 
while WWTP comprises four objectives ($S_I$, $S_S$, $X_I$, and $X_S$) that we maximize. We use V-BREATHE 
for \emph{vector} input. For SOO, we set $\alpha_m = 0.25, \ \forall m \in [1, 2, 3, 4]$, for this 
application, maximizing a simple sum of the COD fractions.

\begin{figure}
    \centering
    \includegraphics[width=0.65\linewidth]{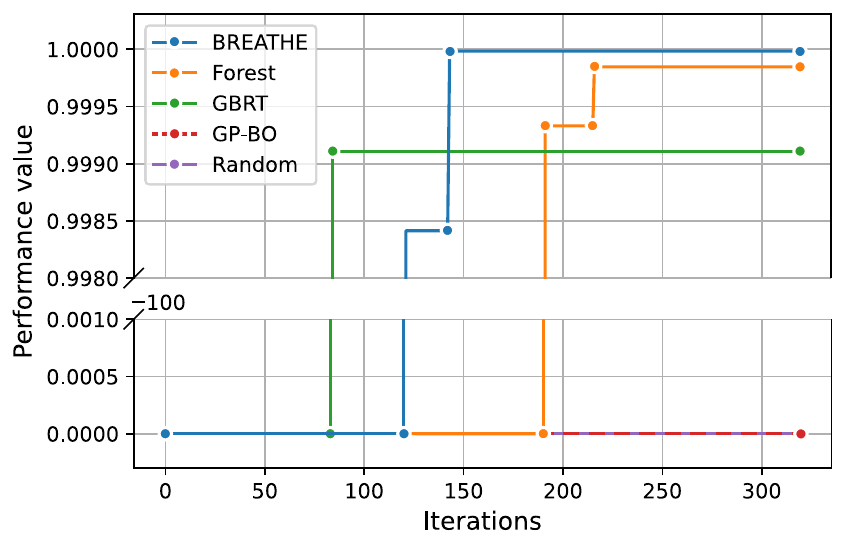}
    \caption{Performance convergence of BREATHE and various baselines on the op-amp application.}
    \label{fig:circuit_soo}
\end{figure}

\begin{figure}
    \centering
    \includegraphics[width=0.65\linewidth]{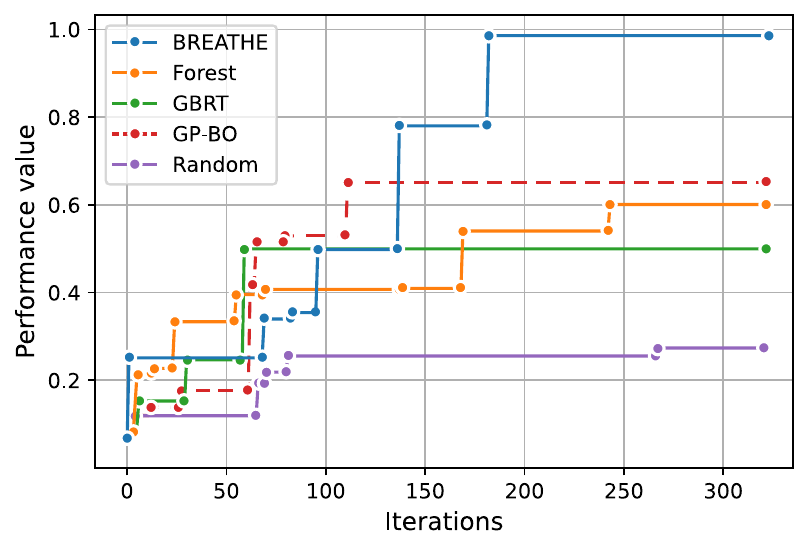}
    \caption{Performance convergence of BREATHE and various baselines on the WWTP application.}
    \label{fig:wwtp_soo}
\end{figure}

Figs.~\ref{fig:circuit_soo} and \ref{fig:wwtp_soo} show the convergence of output performance $P$ for the 
op-amp and WWTP objectives, respectively. BREATHE achieves the highest performance among all methods. 
For the op-amp application, none of the methods can initially find input parameters that satisfy all 
constraints (resulting in the convergence plot to start at $P = -100$). BREATHE, Forest, and GBRT find 
legal inputs and optimize the output performance using these legal data points. However, GP-BO 
and random search are not able to find legal inputs that result in $P > -100$, i.e., $P \in [0, 1]$. 
When there are no constraints, GP-BO achieves the second-highest performance in the WWTP application.

\begin{table}[]
\caption{Selected design parameters of the best-performing points achieved by BREATHE on the op-amp and WWTP applications.}
\centering
\resizebox{\linewidth}{!}{
\begin{tabular}{@{}cccccccccccccc@{}}
\toprule
\multicolumn{14}{c}{\textbf{Op-amp}} \\ \midrule
\multicolumn{2}{c}{$\texttt{VSUPPLY}$} & \multicolumn{2}{c}{$\texttt{VBIAS}$} & $\texttt{IBIAS}$ & $\texttt{CLOAD}$ & $\texttt{RLOAD}$ & $\texttt{C1}$ & $\texttt{C2}$ & $\texttt{L1}$ & $\texttt{L2}$ & $\texttt{L3}$ & $\texttt{L4}$ & $\texttt{L5}$ \\
\multicolumn{2}{c}{2.4 V} & \multicolumn{2}{c}{2.5 V} & 7.0 $\mu$A & 0.12 nF & 10.0 k$\Omega$ & 2.0 pF & 14.0 pF & 4.6 $\mu$m & 6.3 $\mu$m & 2.8 $\mu$m & 1.8 $\mu$m & 3.8 $\mu$m \\ \midrule
\multicolumn{2}{c}{$\texttt{L6}$} & \multicolumn{2}{c}{$\texttt{L7}$} & $\texttt{W1}$ & $\texttt{W2}$ & $\texttt{W3}$ & $\texttt{W4}$ & $\texttt{W5}$ & $\texttt{W6}$ & $\texttt{W7}$ & $\texttt{W8}$ & $\texttt{W9}$ & $\texttt{W10}$ \\
\multicolumn{2}{c}{2.4 $\mu$m} & \multicolumn{2}{c}{5.2 $\mu$m} & 4.4 $\mu$m & 18.4 $\mu$m & 26.3 $\mu$m & 10.4 $\mu$m & 34.3 $\mu$m & 48.4 $\mu$m & 32.4 $\mu$m & 22.3 $\mu$m & 32.3 $\mu$m & 44.4 $\mu$m \\ \midrule
\multicolumn{14}{c}{\textbf{WWTP}} \\ \midrule
$\texttt{S1}$ & $\texttt{S2}$ & $\texttt{RPAX\_T}$ & $\texttt{RPAX\_DO}$ & $\texttt{RPAX\_V}$ & $\texttt{RPOX\_T}$ & $\texttt{RPOX\_DO}$ & $\texttt{RPOX\_V}$ & $\texttt{RAXP\_T}$ & $\texttt{RAXP\_DO}$ & $\texttt{RAXP\_V}$ & $\texttt{ROXP\_T}$ & $\texttt{ROXP\_DO}$ & $\texttt{ROXP\_V}$ \\
1.5 & 0.9 & 10.3\ C & 0.0 mg/L & 3067.0 m$^3$ & 6.2 \textdegree C & 5.0 mg/L & 102.1 m$^3$ & 5.2 \textdegree C & 1.5 mg/L & 100.0 m$^3$ & 13.1 \textdegree C & 3.6 mg/L & 100.0 m$^3$ \\ \bottomrule
\end{tabular}}
\label{tbl:soo_params}
\end{table}

Table~\ref{tbl:soo_params} summarizes the design parameter values for the op-amp and WWTP applications selected by BREATHE. It chooses the input parameters for resistor and capacitor values along with transistor lengths and widths that maximize the output performance. The DC gain of the op-amp is 90.8, its unity gain frequency is 1.58 MHz, and its phase margin is 95.8\textdegree, while incurring 780.2 mW of power. For WWTP, BREATHE chooses a large pre-anoxic reactor (with volume = 3067.0 m$^3$) but much smaller subsequent reactors. Table~\ref{tbl:soo_params} shows other design decisions as well. This design leads to the maximum net COD.

\subsection{Multi-objective Optimization using BREATHE}

\begin{figure}
    \centering
    \includegraphics[width=0.65\linewidth]{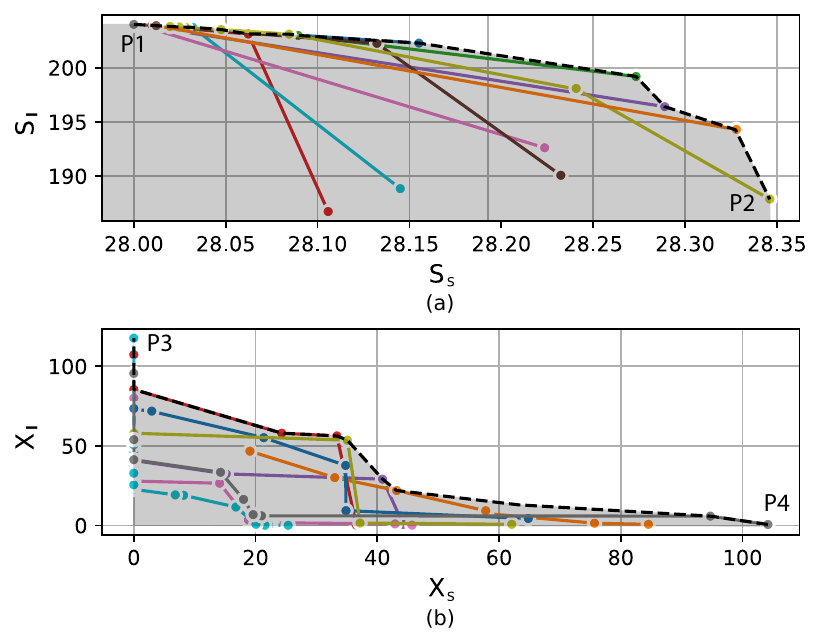}
    \caption{Pareto fronts of (a) $S_I$ vs. $S_S$ and (b) $X_I$ vs. $X_S$ using the BREATHE optimizer 
on the WWTP task. All objectives are plotted in mg/L units.}
    \label{fig:wwtp_pareto}
\end{figure}

We now show the MOO performance of BREATHE on the WWTP application with four output objectives. 
Capturing \emph{non-dominated solutions} on different parts of the Pareto front would require 
contrasting weights for each objective. Thus, we take random samples of the objective weights 
$\alpha_m$. Fig.~\ref{fig:wwtp_pareto} shows the Pareto front while trading off $S_I$ for $S_S$ and 
$X_I$ for $X_S$. \textcolor{black}{These trade-offs are typically studied by domain experts.} Different colors 
correspond to distinct sets of $\alpha_m$'s among 16 random samples. 
We observe that different colors (and thus, weights for the objectives) indeed contribute to unique 
\emph{non-dominated solutions} on the Pareto front (shown by dashed line).

\begin{table}[]
\caption{Selected design parameter values of the four extreme points of the Pareto fronts in Fig.~\ref{fig:wwtp_pareto} for the WWTP application.}
\resizebox{\linewidth}{!}{
\begin{tabular}{@{}l|cccccccccccccc@{}}
\toprule
\multirow{2}{*}{\textbf{Point Name}} & \multicolumn{14}{c}{\textbf{WWTP}} \\ \cmidrule(l){2-15} 
 & $\texttt{S1}$ & $\texttt{S2}$ & $\texttt{RPAX\_T}$ & $\texttt{RPAX\_DO}$ & $\texttt{RPAX\_V}$ & $\texttt{RPOX\_T}$ & $\texttt{RPOX\_DO}$ & $\texttt{RPOX\_V}$ & $\texttt{RAXP\_T}$ & $\texttt{RAXP\_DO}$ & $\texttt{RAXP\_V}$ & $\texttt{ROXP\_T}$ & $\texttt{ROXP\_DO}$ & $\texttt{ROXP\_V}$ \\ \midrule
P1 & 8.0 & 1.0 & 15.0 \textdegree C & 5.0 mg/L & 1458.9 m$^3$ & 15.0 \textdegree C & 0.0 mg/L & 8683.8 m$^3$ & 15.0 \textdegree C & 5.0mg/L & 2450.3 m$^3$ & 15.0 \textdegree C & 0.0 mg/L & 2488.9 m$^3$ \\
P2 & 8.0 & 1.0 & 5.0 \textdegree C & 5.0 mg/L & 11299.7 m$^3$ & 15.0 \textdegree C & 0.0 mg/L & 6899.8 m$^3$ & 15.0 \textdegree C & 0.0 mg/L & 13400.1 m$^3$ & 15.0 \textdegree C & 0.0 mg/L & 14900.0 m$^3$ \\ \midrule
P3 & 1.0 & 1.0 & 15.0 \textdegree C & 0.0 mg/L & 100.0 m$^3$ & 5.0 \textdegree C & 0.0 mg/L & 1518.9 m$^3$ & 5.0 \textdegree C & 0.0 mg/L & 152.9 m$^3$ & 5.0 \textdegree C & 0.0 mg/L & 2779.3 m$^3$ \\
P4 & 8.0 & 1.0 & 5.0 \textdegree C & 3.7 mg/L & 479.9 m$^3$ & 8.8 \textdegree C & 5.0 mg/L & 3504.1 m$^3$ & 14.1 \textdegree C & 5.0 mg/L & 311.1 m$^3$ & 15.0 \textdegree C & 5.0 mg/L & 991.5 m$^3$ \\ \bottomrule
\end{tabular}}
\label{tbl:wwtp_points}
\end{table}

\begin{table}[]
\caption{Output objective values (in mg/L units) for the four extreme points of the Pareto fronts in Fig.~\ref{fig:wwtp_pareto} for the WWTP application.}
\resizebox{0.6\columnwidth}{!}{
\begin{tabular}{@{}l@{\hskip 0.4in}|@{\hskip 0.4in}c@{\hskip 0.4in}c@{\hskip 0.4in}c@{\hskip 0.4in}c@{}}
\toprule
\multirow{2}{*}{\textbf{Point Name}} & \multicolumn{4}{c}{\textbf{Output Objectives}} \\ \cmidrule(l){2-5} 
 & $S_I$ & $S_S$ & $X_I$ & $X_S$ \\ \midrule
P1 & \textbf{204.01} & 28.00 & 36.34 & 0.01 \\
P2 & 193.93 & \textbf{28.45} & 13.37 & 0.27 \\ \midrule 
P3 & 204.00 & 27.94 & \textbf{117.64} & 0.01 \\
P4 & 1.60 & 19.52 & 0.58 & \textbf{104.17} \\ \bottomrule
\end{tabular}}
\label{tbl:wwtp_points_objectives}
\end{table}

Table~\ref{tbl:wwtp_points} summarizes the design choices of the four plants that correspond to the four extrema of the
two Pareto fronts (P1-P4) in Fig.~\ref{fig:wwtp_pareto}. P1 corresponds to the plant with the highest achieved $S_I$ in
the effluent, while P4 corresponds to the plant with the highest achieved $X_S$ in the effluent.
Table~\ref{tbl:wwtp_points_objectives} shows the output objective values for the four design points. When we maximize
one objective, other objective values are much lower. This implies that there is a trade-off when maximizing all objectives that BREATHE considers by presenting a 
Pareto front. 

\begin{figure}
    \centering
    \includegraphics[width=0.65\linewidth]{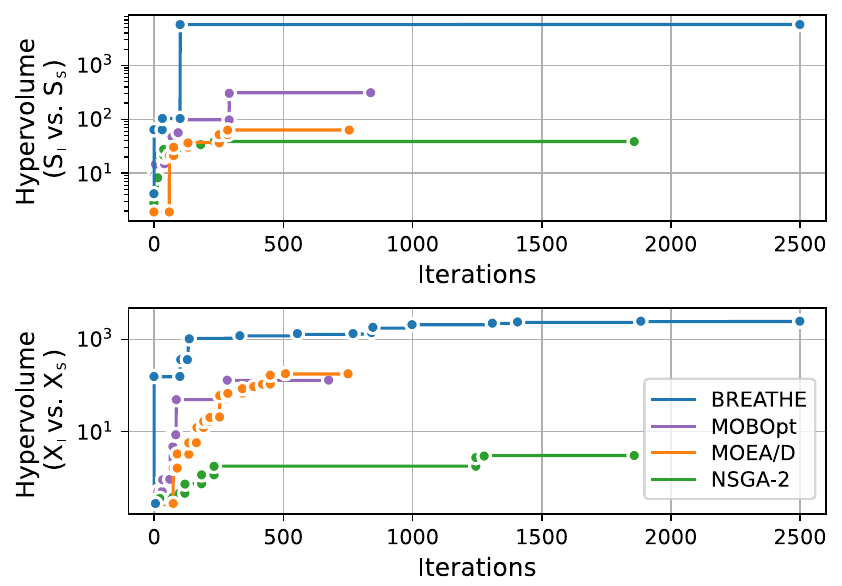}
    \caption{Convergence of hypervolume (in mg$^2$/L$^2$ units) of (a) $S_I$ vs. $S_S$ and (b) $X_I$ 
vs. $X_S$ Pareto fronts with the number of iterations when running BREATHE and other MOO baselines on 
the WWTP task.}
    \label{fig:wwtp_hypervolume}
\end{figure}

We train independent surrogate models for each selection of weights and run the BREATHE optimization 
pipeline in parallel. We observe that parallel runs outperform sequential operations of the BREATHE 
algorithm (where we iteratively update the surrogate model on each new dataset with the new set of 
objective weights). We hypothesize that the independent parallel runs result in a higher variance in the 
internal representations of the meta-model as it covers a larger fraction of the design space. 

Hypervolume is a measure of the solution quality in MOO problems. We derive it by measuring the size of 
the \emph{dominated} portion of the design space~\cite{hypervolume}, i.e., the area under the Pareto front 
above zero. In Fig.~\ref{fig:wwtp_pareto}, we shade the area that we use to compute the hypervolume in 
grey. Fig.~\ref{fig:wwtp_hypervolume} shows the convergence of hypervolume with the number of algorithm 
iterations for BREATHE and baseline methods. Numerous iterations of BREATHE correspond to multiple runs 
(till convergence) using different randomly sampled weights. Since we need to maximize all objectives, we 
must also maximize the resultant hypervolume. BREATHE achieves a considerably higher hypervolume relative 
to baselines with fewer queried samples. More concretely, BREATHE achieves 21.9$\times$ higher hypervolume relative to MOBOpt in the $S_I$ vs. $S_S$ trade-off and 20.1$\times$ higher hypervolume in the $X_I$ vs. $X_S$ trade-off (more details in Section~\ref{sec:discussion}).

\begin{figure}
    \centering
    \includegraphics[width=\linewidth]{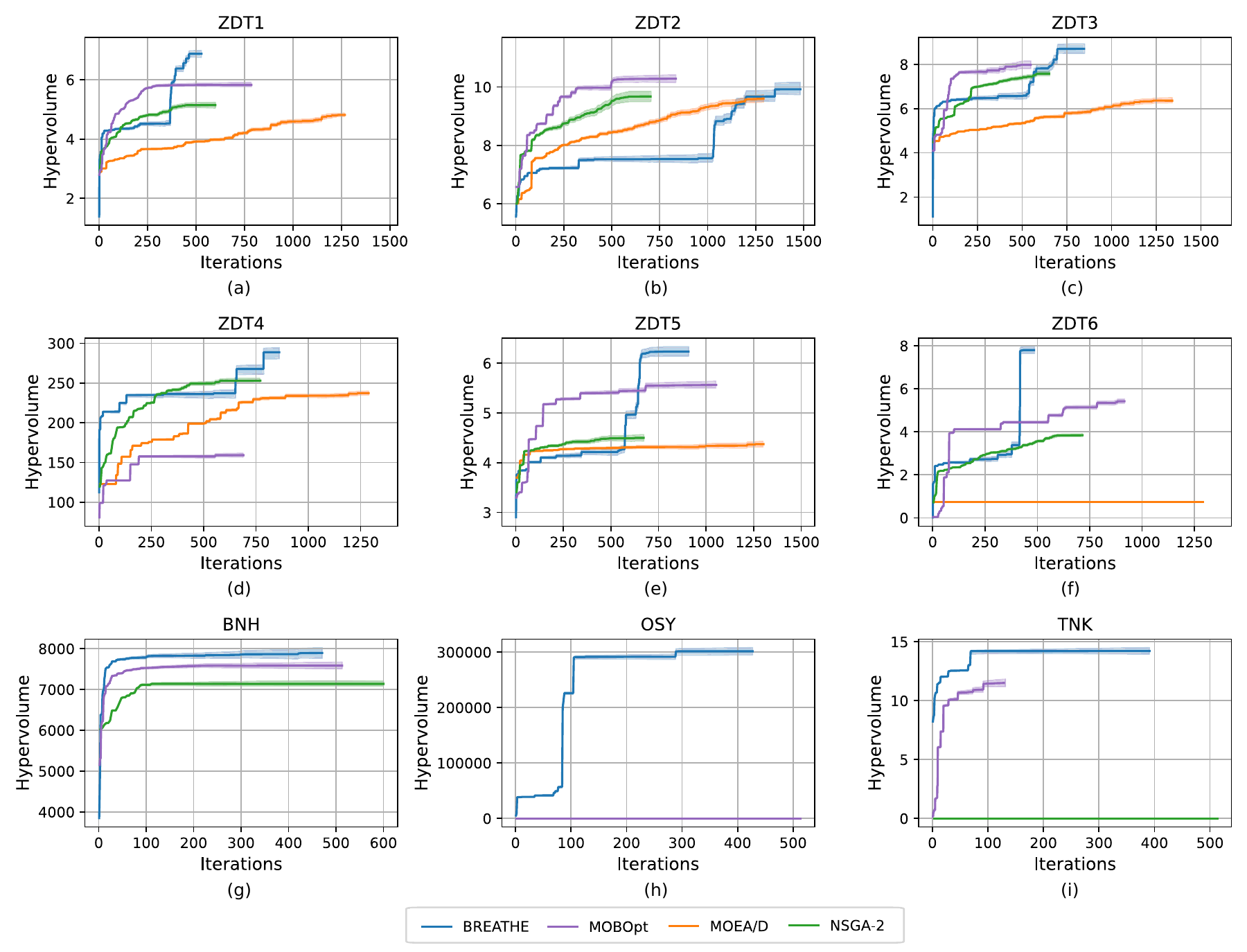}
    \caption{Convergence of hypervolume, when comparing BREATHE against baselines, on the (a-f) ZDT1-6 problem suite, (g) BNH, (h) OSY, and (i) TNK benchmarks. All methods were executed until convergence. For all plots, we show the mean and one standard error of the mean over five replications.}
    \label{fig:benchmarks}
\end{figure}

Fig.~\ref{fig:benchmarks} shows the convergence of hypervolume with the number of algorithm iterations for BREATHE and baseline methods on benchmark applications. Here, we show the hypervolume that is dominated by the provided set of solutions with respect to a reference point (set by the maximum possible value of each output objective)~\cite{hypervolume_algo}. We observe that BREATHE outperforms baselines on most benchmarks (except ZDT2). MOEA/D does not support constraints and is therefore not plotted for BNH, OSY, and TNK tasks. Further, for the OSY task, NSGA-2 and MOBOpt could not find any legal inputs (that satisfy all constraints), resulting in the hypervolume being zero. 

\begin{figure}
    \centering
    \includegraphics[width=\linewidth]{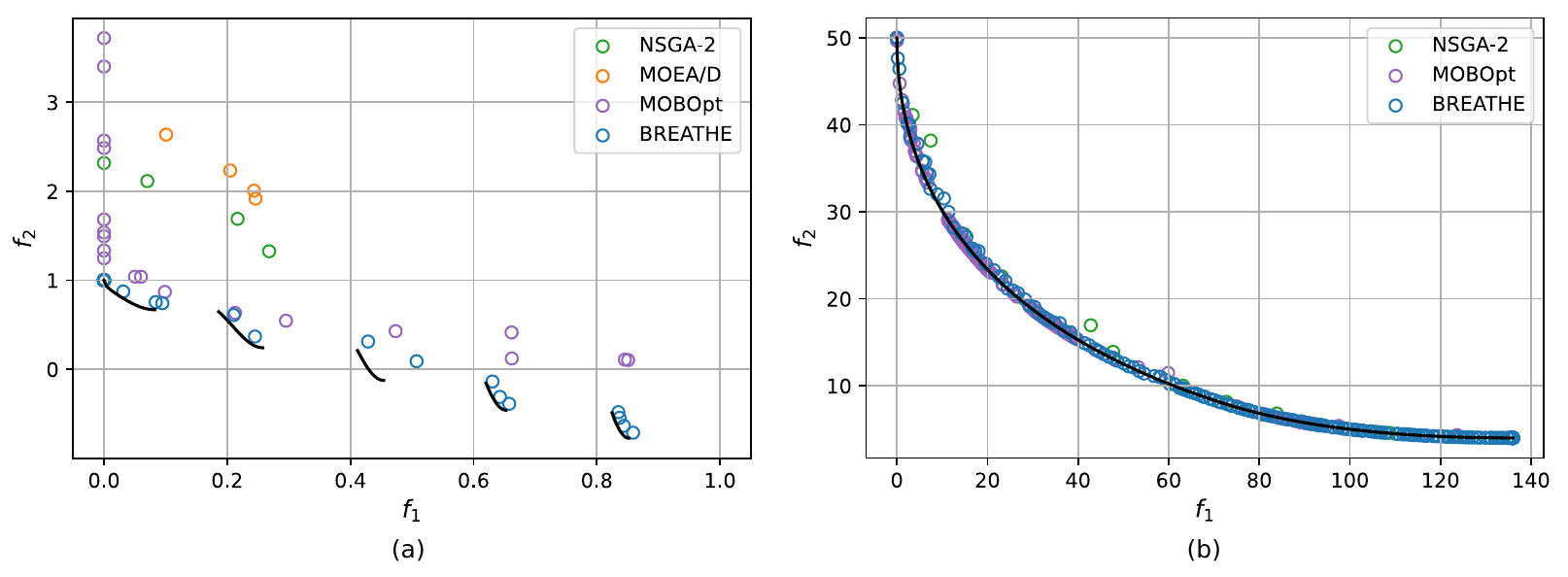}
    \caption{Pareto fronts obtained by various MOO methods on the (a) ZDT3 and (b) BNH tasks.}
    \label{fig:pareto_zdt3_bnh}
\end{figure}

Fig.~\ref{fig:pareto_zdt3_bnh} shows the obtained Pareto fronts for the ZDT3 and BNH tasks. ZDT3 has a disjoint and non-convex Pareto front. Even though BREATHE uses a convex combination scalarization function 
[see Eq.~(\ref{eq:perf_moo})], which has been shown to not perform well for non-convex Pareto 
fronts~\cite{scalarizing}, it is able to obtain non-dominated solutions close to ZDT3's Pareto front due to 
multiple random cold restarts in the optimization loop. In Fig.~\ref{fig:pareto_zdt3_bnh}(b), we show that BREATHE is able to achieve a denser set of non-dominated solutions on the Pareto front relative to baselines.

\begin{figure}
    \centering
    \includegraphics[width=0.65\linewidth]{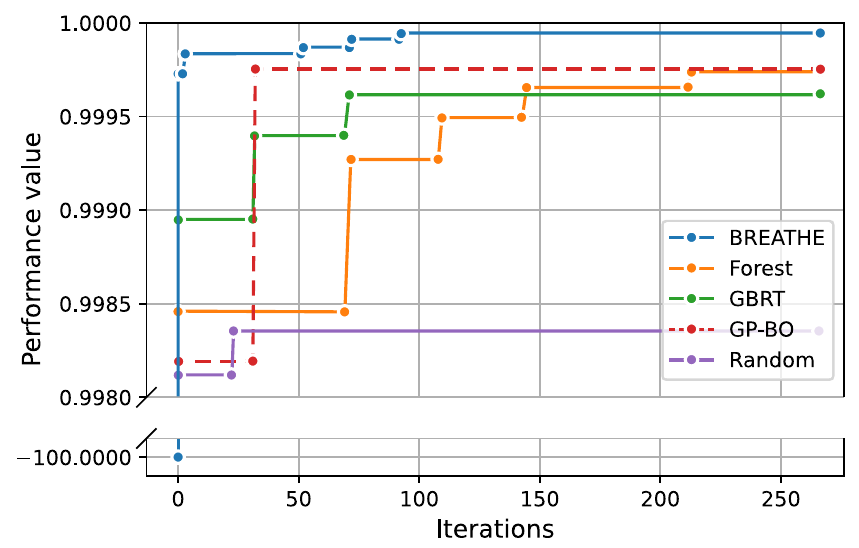}
    \caption{Performance convergence of BREATHE on the smart home application.}
    \label{fig:smart_home_soo}
\end{figure}

\subsection{Searching for Optimal Graphs using BREATHE}

We now run the G-BREATHE algorithm, as described in Section~\ref{sec:gbreathe}, for graph optimization. 
This implies searching for novel graphs along with node and edge weights that maximize the output 
performance measure $P$. Fig.~\ref{fig:smart_home_soo} shows the convergence of $P$ with the number of 
iterations on the smart home task. This task has only one objective: minimization of the number of attacks. BREATHE outperforms all baselines. Even though randomly generated graphs may be legal in terms of the graph architecture, they may not honor all the constraints (for example, the number of windows should be greater than three). BREATHE and the baselines quickly find legal graphs (with $P \in [0, 1]$). However, not being able to smartly search the graph architecture space limits the baselines 
from reaching the highest-achieved performance by BREATHE. 

\begin{figure}
    \centering
    \includegraphics[width=0.65\linewidth]{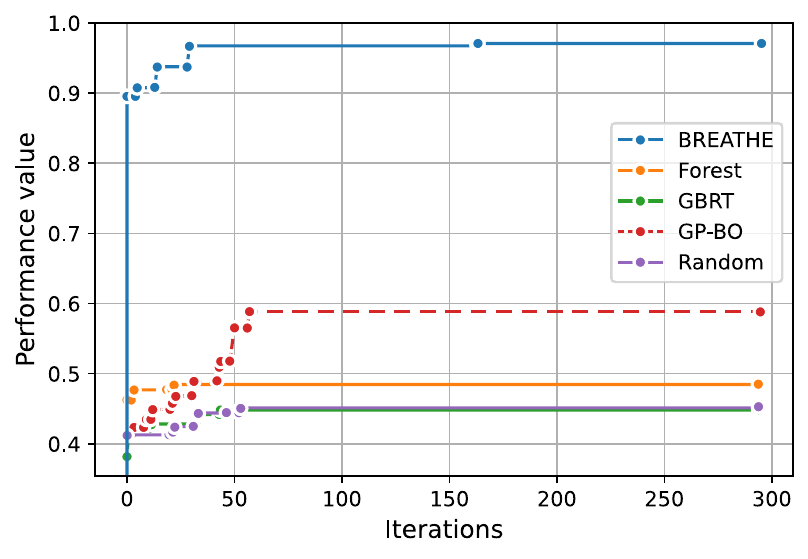}
    \caption{Performance convergence of BREATHE on the network application.}
    \label{fig:network_soo}
\end{figure}

Fig.~\ref{fig:network_soo} shows the performance convergence for network optimization. This task has two 
objectives: maximization of the average bandwidth and minimization of overall network operation cost. 
The weights for the two objectives for the calculation of $P$ are 0.4 and 0.6, respectively. We choose these 
weights to attribute more importance to the network operation cost. A user can choose any set of weights that form a convex combination (as in 
Eq.~(\ref{eq:graph_constrained_moo})). Again, BREATHE outperforms baselines by achieving 64.9\% higher performance than the 
next-best baselines, i.e., GP-BO (more details in Section~\ref{sec:discussion}).

\begin{figure}
    \centering
    \includegraphics[width=0.7\linewidth]{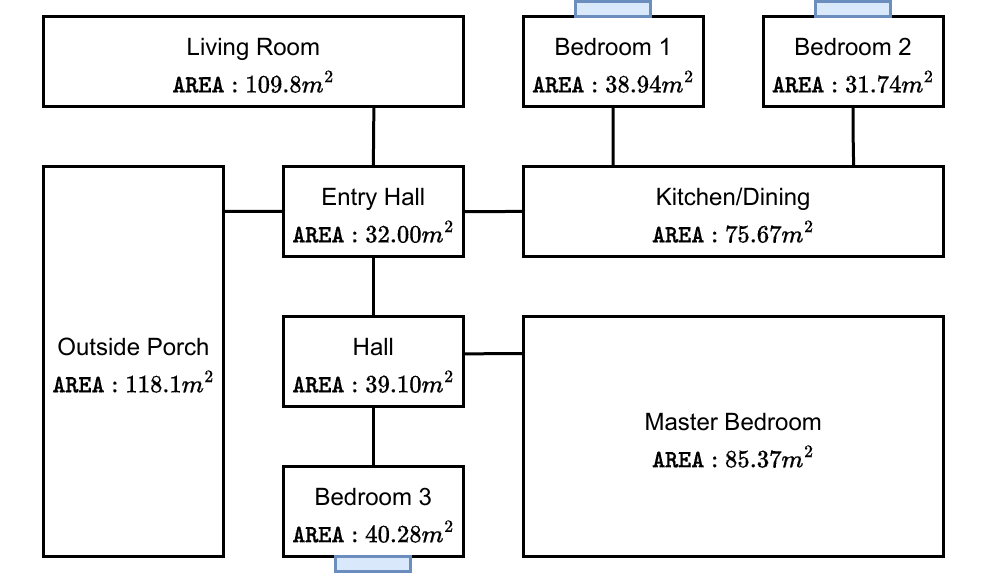}
    \caption{Best-performing smart home design that minimizes the number of cyber-physical attacks as per G-BREATHE.}
    \label{fig:smart_home}
\end{figure}

Fig.~\ref{fig:smart_home} shows the obtained smart home design after running the BREATHE algorithm. All bedrooms have a window (shown by a blue box). Bedroom 1 has a WiFi modem and a smart thermostat. The entry hall contains an entry sensor, a smart door camera, and a smart lock at the door connecting to the outside porch. The kitchen/dining area has a home assistant and the 
master bedroom has a network gateway. Placing the gateway in a different room than the WiFi modem reduces the risk of attacks. Nevertheless, this design is prone to 65 physical (break-ins from the outside door or windows) and 92 cyber attacks (DDoS attacks affecting the gateway and, subsequently, the IoT devices). These correspond to different permutations of attacks one can perform. However, this is the least number of cyber-physical attacks achieved by the BREATHE algorithm.

\begin{figure}
    \centering
    \includegraphics[width=0.5\linewidth]{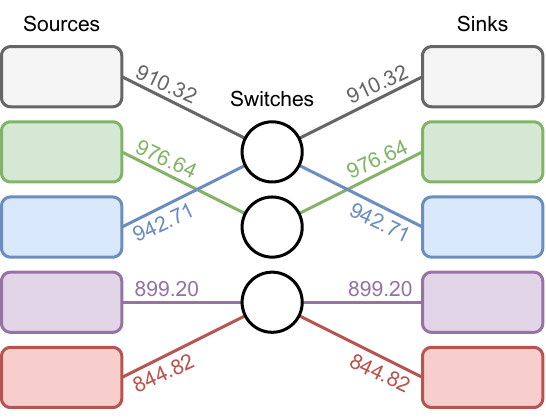}
    \caption{Best-performing network architecture that maximizes performance as per G-BREATHE. An annotation on a connection represents the bandwidth of that connection in MB/s.}
    \label{fig:network}
\end{figure}

Multiple network configurations lead to the same net performance value. For example, Fig.~\ref {fig:network} shows one such best-performing network architecture. The cost of operation for the network is \$35,457.37 and the average bandwidth is 
914.74 MB/s. The network only uses three switches to minimize the base cost of setting them up. Two switches connect to two data source/sink pairs, while one switch connects to only one pair. We label these pairs and their corresponding connections in unique colors.

\subsection{Ablation Analysis}

\begin{figure}
    \centering
    \includegraphics[width=0.65\linewidth]{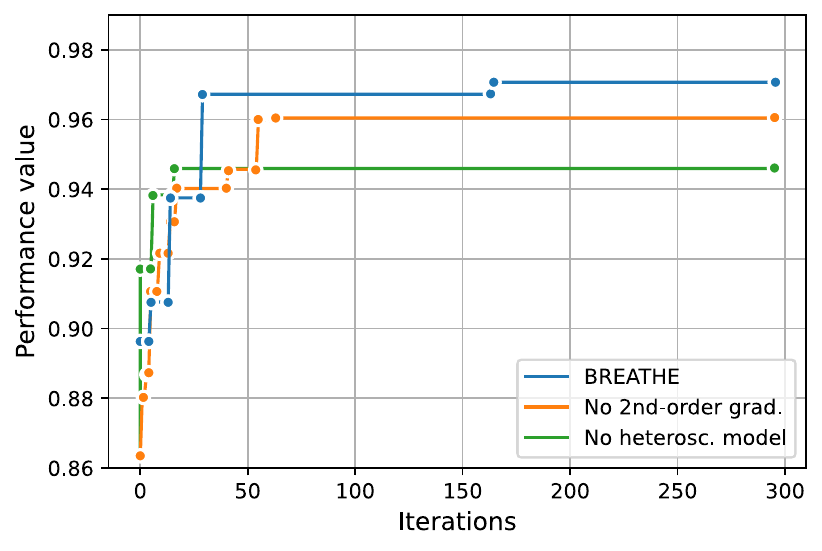}
    \caption{Ablation analysis of BREATHE on the network application.}
    \label{fig:network_ablation}
\end{figure}

We now present an ablation analysis of our proposed optimizer. Fig.~\ref{fig:network_ablation} compares 
the performance convergence of BREATHE with that of its ablated versions. First, we remove second-order 
gradients ($\nabla^2_x$UCB) and implement GOBI with first-order gradients ($\nabla_x$UCB) instead. Second, 
we remove the NPN model, which models the aleatoric uncertainty (see Section~\ref{sec:vbreathe}). We can 
see that these changes result in poorer converged performance relative to the proposed BREATHE algorithm. 

\textcolor{black}{We now ablate the effect of the proposed \emph{legality-forcing} method on benchmark applications with constraints. We present the results in Table~\ref{tbl:legality_forcing_ablation}. We observe that legality-forcing is crucial for constrained optimization. Without it, GOBI could result in inputs that are not legal.}

\begin{table}[]
\caption{\textcolor{black}{Effect of legality-forcing on constrained optimization using BREATHE.}}
\resizebox{0.45\columnwidth}{!}{
\begin{tabular}{@{}l@{\hskip 0.2in}|@{\hskip 0.2in}c@{\hskip 0.2in}c@{\hskip 0.2in}c@{}}
\toprule
\multirow{2}{*}{\textbf{Method}} & \multicolumn{3}{c}{\textbf{Application}} \\ \cmidrule(l){2-4} 
                                 & BNH          & OSY           & TNK       \\ \midrule
V-BREATHE                        & 7894.4       & 301429.9      & 14.2      \\
w/o legality-forcing             & 6933.5       & -             & 7.8       \\ \bottomrule
\end{tabular}}
\label{tbl:legality_forcing_ablation}
\end{table}

\subsection{Scalability Tests}
\label{sec:scalability}

\begin{figure}
    \centering
    \includegraphics[width=0.65\linewidth]{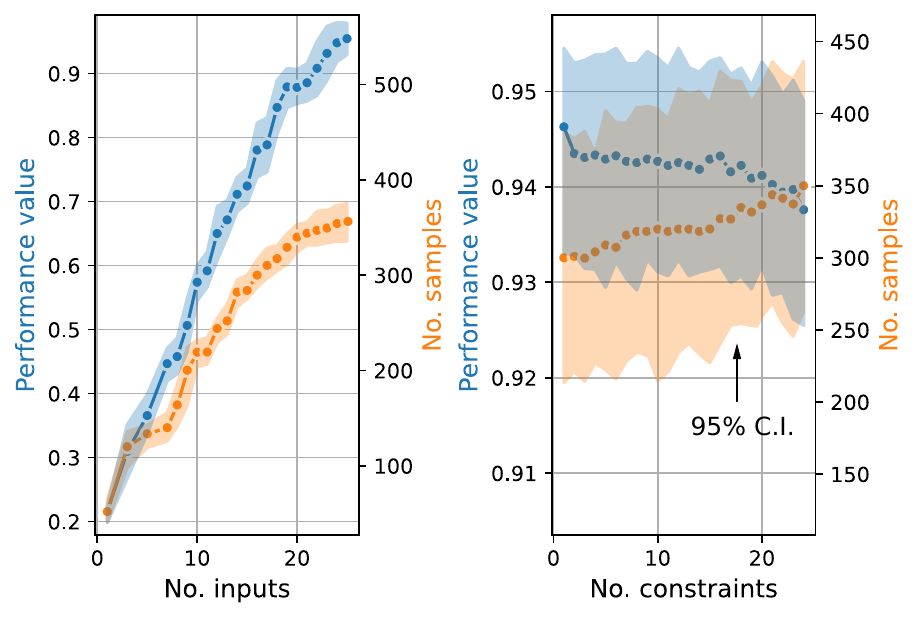}
    \caption{Effect of the number of (a) inputs and (b) constraints on the highest achieved performance 
value and the number of samples required to achieve it. BREATHE was executed on the SO-PWM task. Plotted 
with 95\% confidence intervals.}
    \label{fig:sopwm_scalability}
\end{figure}

We now test how scalable BREATHE is with the dimensionality of the optimization problem. Hence, we plot 
how the maximum achieved performance and the number of samples required to achieve that performance scale 
with an increasing number of inputs and constraints. \textcolor{black}{Here, the number of samples
corresponds to the total number of queries to the simulator. These include the seed samples required to initialize and train the surrogate model.} To calculate the performance ($P$) on the SO-PWM task, we use $\alpha_1 = \alpha_2 = 0.5$ to give equal weight to each objective. We show these plots in 
Fig.~\ref{fig:sopwm_scalability}. First, we note that the maximum achieved performance scales linearly 
with the number of inputs and constraints. However, with increasing constraints, the maximum achieved 
performance value slowly decreases. Second, sample complexity scales sublinearly with the number of 
inputs and linearly with constraints.

\section{Discussions and Limitations}
\label{sec:discussion}

\begin{table}[]
\caption{Summary of experimental results for various optimization applications. Best-achieved performance is reported
for SOO tasks and the highest achieved hypervolume is reported for the MOO tasks. $^*$The graphical version of the baseline was implemented as described in Section~\ref{sec:baselines}.}
\resizebox{0.65\columnwidth}{!}{
\begin{tabular}{@{}lccccc@{}}
\toprule
\multicolumn{6}{c}{\textbf{SOO (Vector Optimization)}} \\ \midrule
\multicolumn{1}{l|}{\textbf{Application}} & \multicolumn{1}{l}{\textbf{Random}} & \textbf{GP-BO} & \textbf{GBRT} & \textbf{Forest} & \textbf{V-BREATHE} \\ \midrule
\multicolumn{1}{l|}{Op-amp} & -100.0 & -100.0 & 0.9991 & 0.9997 & \textbf{0.9999} \\ [1mm]
\multicolumn{1}{l|}{WWTP} & 0.2735 & 0.6517 & 0.4996 & 0.6006 & \textbf{0.9858} \\ \midrule
\multicolumn{6}{c}{\textbf{SOO (Graph Optimization)}} \\ \midrule
\multicolumn{1}{l|}{\textbf{Application}} & \multicolumn{1}{l}{\textbf{Random$^*$}} & \textbf{GP-BO$^*$} & \textbf{GBRT$^*$} & \textbf{Forest$^*$} & \textbf{G-BREATHE} \\ \midrule
\multicolumn{1}{l|}{Smart Home} & 0.9983 & 0.9997 & 0.9996 & 0.9997 & \textbf{0.9999} \\ [1mm]
\multicolumn{1}{l|}{Network} & 0.4512 & 0.5885 & 0.4482 & 0.4846 & \textbf{0.9707} \\ \midrule
\multicolumn{6}{c}{\textbf{MOO (Vector Optimization)}} \\ \midrule
\multicolumn{1}{l|}{\textbf{Application}} & \textbf{NSGA-2} & \multicolumn{2}{c}{\textbf{MOEA/D}} & \textbf{MOBOpt} & \textbf{V-BREATHE} \\ \midrule
\multicolumn{1}{l|}{WWTP ($S_I$ vs. $S_S$)} & 39.4 & \multicolumn{2}{c}{67.6} & 263.9 & \textbf{5781.3} \\ [1mm]
\multicolumn{1}{l|}{WWTP ($X_I$ vs. $X_S$)} & 3.2 & \multicolumn{2}{c}{182.3} & 150.2 & \textbf{3024.1} \\ \midrule
\multicolumn{1}{l|}{ZDT1} & 5.1 & \multicolumn{2}{c}{4.8} & 5.9 & \textbf{7.1} \\
\multicolumn{1}{l|}{ZDT2} & 9.7 & \multicolumn{2}{c}{9.6} & \textbf{10.3} & 9.9 \\
\multicolumn{1}{l|}{ZDT3} & 7.6 & \multicolumn{2}{c}{6.4} & 8.0 & \textbf{8.5} \\
\multicolumn{1}{l|}{ZDT4} & 250.7 & \multicolumn{2}{c}{238.6} & 159.1 & \textbf{282.4} \\
\multicolumn{1}{l|}{ZDT5} & 4.5 & \multicolumn{2}{c}{4.4} & 5.6 & \textbf{6.2} \\
\multicolumn{1}{l|}{ZDT6} & 3.8 & \multicolumn{2}{c}{0.7} & 5.4 & \textbf{7.8} \\
\multicolumn{1}{l|}{BNH} & 7137.2 & \multicolumn{2}{c}{-} & 7635.2 & \textbf{7894.4} \\
\multicolumn{1}{l|}{OSY} & - & \multicolumn{2}{c}{-} & - & \textbf{301429.9} \\
\multicolumn{1}{l|}{TNK} & - & \multicolumn{2}{c}{-} & 11.7 & \textbf{14.2} \\ \bottomrule
\end{tabular}}
\label{tbl:results_summary}
\end{table}

Table~\ref{tbl:results_summary} summarizes the results presented in Section~\ref{sec:results}. The proposed 
framework outperforms baselines on various applications. In SOO, BREATHE achieves 64.1\% higher performance 
than the next-best baseline, i.e., Forest on the WWTP application. G-BREATHE achieves 64.9\% higher performance than the next-best baseline, i.e., a graphical version of GP-BO, on the network application. In MOO, BREATHE achieves up to 21.9$\times$ higher 
hypervolume relative to the next-best baseline, namely, MOBOpt (in the $S_I$ vs. $S_S$ trade-off). BREATHE also outperforms baselines on standard MOO benchmark applications\textcolor{black}{, except ZDT2. ZDT2 and ZDT6 are MOO problems with non-convex Pareto fronts. However, ZDT6 has only 10 inputs, while ZDT2 has 30 inputs. This leads us to believe that BREATHE may not always outperform the baselines when the optimization problem is non-convex and has high input dimensionality.}

Although the surrogate model that undergirds the V-BREATHE framework is similar to that of BOSHNAS~\cite{flexibert}, many novelties are proposed in this work. These include \emph{legality-forcing} of gradients to allow queries to adhere to input constraints, \emph{penalization} for output constraint violations, and support for multi-objective optimization \textcolor{black}{(although, we leave the exploration of more complex constraint management methods, like probability of feasibility~\cite{sohst2022surrogate}, to future work)}. Moreover, BOSHNAS only works with vector input, while G-BREATHE optimizes for graph architectures. G-BREATHE does not convert the graph-based problem into a vector-based problem. Instead, it directly works on graphical input. It not only optimizes for the node/edge weight values (that could, in principle, be reduced to a vector optimization problem) but also searches for the best-performing graph architecture (node connections, i.e., new edges). Graph architecture optimization has not been implemented by any previously-proposed surrogate-based optimization method, to the best of our knowledge.

In the demonstrated results, we show that 
different extrema on the Pareto front result in designs that maximize one objective at the cost of others and BREATHE outputs 
all such extrema. G-BREATHE achieves high performance values in graph search, resulting in designs that optimize output
objectives while honoring user-defined constraints. Unlike previous works, BREATHE is a unified framework that supports
sample-efficient optimization in different input spaces (vector- or graph-based) and user-defined constraints. This work
shows the applicability of BREATHE to diverse optimization problems and explores the novel domain of \emph{graph optimization} 
for generic applications (beyond NAS). BREATHE also leverages an actively trained heteroscedastic model to minimize sample complexity. 

\textcolor{black}{In this work, we tested G-BREATHE on graphical spaces with up to 25 nodes. The proposed G-BREATHE framework is applicable to larger graphs as well. However, case studies involving larger and more complex graphs would require domain expertise and modifications to the optimization method for better-posed search. This is because larger graphs exponentially increase the size of the design space. Exploring such cases is part of future work.}

BREATHE has several limitations. It only supports optimization of an application where the simulator gives an output 
for any legal queried input. However, in some cases, the queried simulator could result in erroneous output. BREATHE
does not detect such outputs based on the distributions learned from other input/output pairs. Detecting such pairs
falls under the scope of adversarial attack detection~\cite{adv_attack_detection} and label noise 
detection~\cite{confident_learning, ctrl}. Moreover, its surrogate model does not work with partially-specified inputs. 

\section{Conclusion}
\label{sec:conclusion}

In this work, we presented BREATHE, a vector-space and graph-space optimizer that \emph{efficiently} 
searches the design space for constrained single- or multi-objective optimization applications. It leverages 
second-order gradients and actively trains a heteroscedastic model by iteratively querying an expensive 
simulator. BREATHE outperforms the next-best baseline, Forest, with up to 64.1\% higher performance. 
G-BREATHE is an optimizer that efficiently searches for graphical architectures along with node and edge 
weights. It outperforms the next-best baseline, a graphical version of GP-BO, with up to 64.9\% higher performance. Further, we leverage BREATHE for multi-objective optimization where it achieves up to 21.9$\times$ 
higher hypervolume than a state-of-the-art baseline, MOBOpt.

\bibliographystyle{ACM-Reference-Format}
\bibliography{biblio}

\end{document}